\documentclass[runningheads]{llncs}
\usepackage[T1]{fontenc}
\usepackage{graphicx}
\usepackage{multirow}
\usepackage{url}

\begin{document}
\title{FieldWorkArena: Agentic AI Benchmark for Real Field Work Tasks}
%
%
\author{
  Jun Takahashi\inst{1} \and
  Atsunori Moteki\inst{1} \and
  Akiyoshi Uchida\inst{1} \and
  Shoichi Masui\inst{5} \and
  Fan Yang\inst{2} \and
  Kanji Uchino\inst{2} \and
  Yueqi Song\inst{3} \and
  Yonatan Bisk\inst{3} \and
  Graham Neubig\inst{3} \and
  Ikuo Kusajima\inst{1} \and
  Yasuto Watanabe\inst{1} \and
  Hiroyuki Ishida\inst{1} \and
  Koki Nakagawa\inst{4}\thanks{Intern} \and
  Shan Jiang\inst{1}
  }
\authorrunning{J. Takahashi et al.}
%
\institute{Fujitsu Limited, Kanagawa, Japan \and
Fujitsu Research of America, CA, USA \and
Carnegie Mellon University, Pittsburgh, USA \and
Master's Student, The University of Tokyo, Tokyo, Japan \and
Agent Research Collective, Kanagawa, Japan \\
\email{takahashi\_jun@fujitsu.com}
}

\maketitle              

\begin{abstract}
This paper introduces \textit{FieldWorkArena}, a benchmark for agentic AI targeting real-world field work.
With the recent increase in demand for agentic AI, they are built to detect and document safety hazards, procedural violations, and other critical incidents across real-world manufacturing and retail environments.
Whereas most agentic AI benchmarks focus on performance in simulated or digital environments, our work addresses the fundamental challenge of evaluating agents in the real-world.
In this paper, we improve the evaluation function from previous methods to assess the performance of agentic AI in diverse real-world tasks.
Our dataset comprises on-site captured images/videos in factories, warehouses and retails.
Tasks were meticulously developed through interviews with site workers and managers.
Evaluation results confirmed that performance evaluation considering the characteristics of Multimodal LLM (MLLM) such as GPT-4o is feasible.
Furthermore, this study identifies both the effectiveness and limitations of the proposed new evaluation methodology.
The complete dataset and evaluation program are publicly accessible on the website\footnote{\url{https://en-documents.research.global.fujitsu.com/fieldworkarena/} \label{foot:fieldworkarena}}.
\end{abstract}

%
%

\section{Introduction}

To address labor shortages and improve efficiency in manufacturing, logistics, and other fields, there is growing interest in using agentic AI for on-site support. 
In this paper, agentic AI refers to autonomous intelligent systems capable of executing specific tasks without human intervention.
The introduction of agentic AI in the workplace is expected to enable functions such as extracting necessary knowledge from documents, analyzing real time on-site conditions using sensors like cameras, and notifying relevant members of required information.
However, technical challenges in the field implementation of agentic AI include addressing diverse and multi-stage tasks and adapting to complex and dynamic on-site environments. 
To assess the effectiveness of agentic AI in addressing these challenges, appropriate benchmarks are necessary to objectively evaluate their performance. 
However, existing agentic AI benchmarks \cite{zhou2024webarena,koh2024visualwebarena,drouin2024workarena} focus on performance in simulated or digital environments, making it difficult to verify their versatility and robustness in solving tasks in actual field environments. 
Furthermore, benchmarks that support multimodal inputs such as images, videos and PDF documents obtained from the field are limited.

\begin{figure}[htbp]
  \centering
  \includegraphics[width=\textwidth]{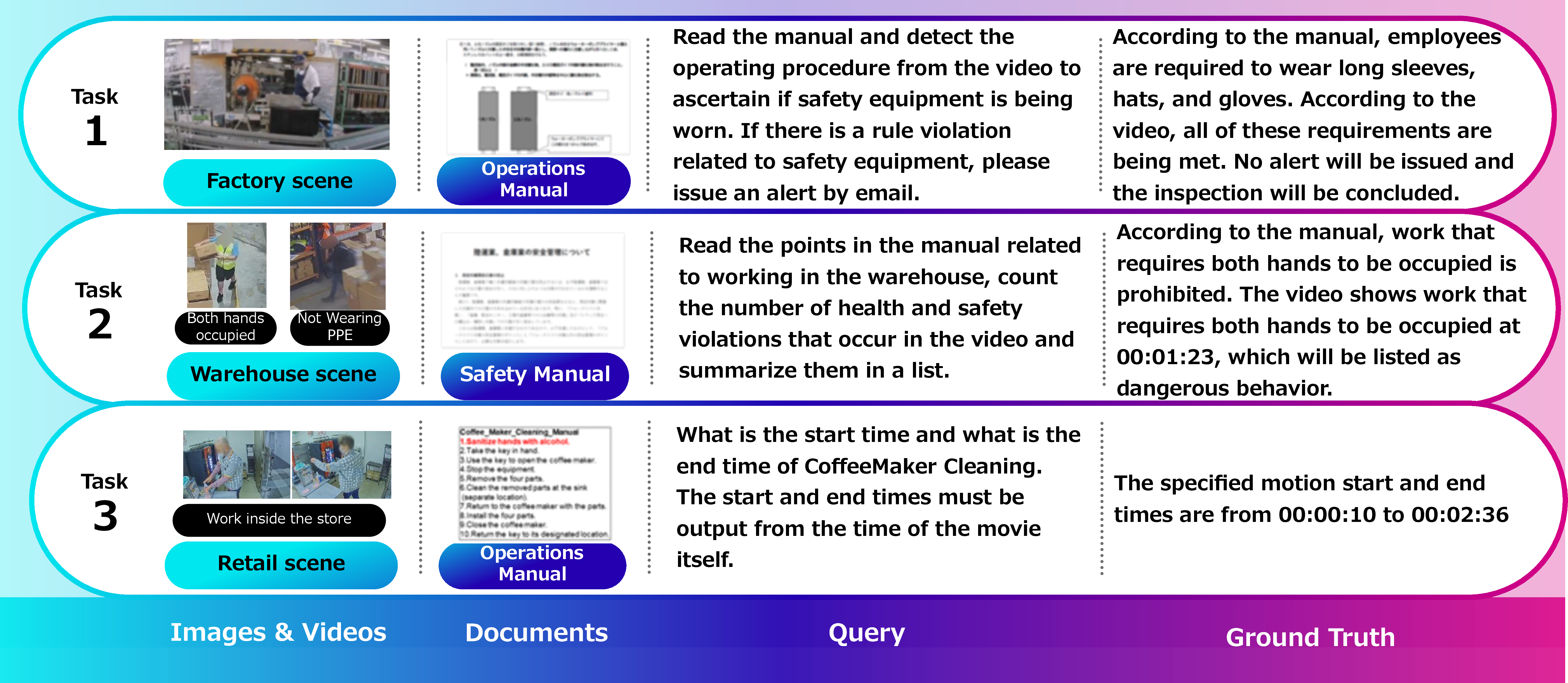}
  \caption{Example of \textit{FieldWorkArena} dataset which includes images and videos taken on site, documents, queries, and ground truth.}
  \label{fig:teaser}
\end{figure}

To address this, we propose an agentic AI benchmark suite \textit{FieldWorkArena}, a benchmark suite aimed at promoting the introduction of agentic AI in performing monitoring tasks in fieldwork environments. 
The dataset configuration is shown in Figure \ref{fig:teaser}. 
FieldWorkArena includes over 400 types of data (images, videos, work manuals) and approximately 900 field-specific queries from three real-world field scenarios (factories, warehouses, and retail), along with manually annotated ground truth data.
By defining the functions that agentic AI should process for each category as an action space, the system can comprehensively handle various types of on-site tasks. 
This enables users to train existing Multimodal LLMs (MLLMs) or agentic AI under development, as well as perform quantitative evaluation and comparison.

\section{FieldWorkArena Data Collection}
\label{sec:impl}

This section describes the types of image and video data, preprocessing methods, task generation, and annotation guidelines for \textit{FieldWorkArena}.
The dataset is available on Hugging Face, but user registration is required to download the data from the web page.
The data is provided under a custom license (detailed on the project website) that restricts its use to non-commercial purposes and prohibits redistribution.
This approach is rooted in our data governance principles.
Although the dataset was released after following procedures based on our internal guidelines for protecting workers' personal information, these strict licensing terms are in place to further ensure the privacy and rights of the individuals depicted.
The evaluation program, in contrast, is separately licensed under the Apache-2.0.

\subsection{Input Data Acquisition}
\label{sec:impl_acquisition}

Data was collected from factories, warehouses, and retail environments, including video footage, work manuals, and safety regulations.
Image and video data have been obtained with consent from all persons appearing in the footage for public disclosure on the web, and faces and necessary areas have been blurred to prevent identification of individuals.

Factory: The images and videos for the factory scenario in \textit{FieldWorkArena} were captured by 11 cameras installed in a communications equipment manufacturing facility.
The footage includes close-up shots of assembly processes for server PCs and telephone equipment, soldering tank cleaning operations, as well as wide-angle shots of verification work on semiconductor production lines.
These 16 types of videos were divided into evaluation videos ranging from several seconds to several minutes in length, resulting in 64 evaluation videos. Additionally, 31 evaluation images are extracted from the videos.
The resolution ranges from 960x540 to 1280x720, and the frame rate is 15 to 30 FPS.
Two types of documents were prepared for workflow extraction: a task list and a task manual. The task list describes the work procedures captured by the camera, labeled with labels and summaries. The task manual includes instructions such as how to wear gloves and how to clean equipment. Both were modified as appropriate by removing proper nouns from the actual processes and manuals used.

Warehouse: Video footage and images of the warehouse are captured using 8 fixed-position cameras installed in a warehouse that handles electronic devices. 
Each warehouse operation, that is, receiving, acceptance, sorting, storage, and shipping is filmed in 6 types, and the footage is divided into 25 videos and 61 images.
The resolution is 1920x1080, and the frame rate is 15 to 25 FPS.
Since the documents related to warehouse operations cannot be made public, we have modified and adapted manuals related to general warehouse operations and forklift usage from other sources for use in this project.

Retail: Video footage and images are captured in a real convenience store and a supermarket in Japan.
Unlike other cases, it is difficult to obtain the consent from all customers appearing in the footage for public disclosure on the web.
Therefore, for retail tasks, while the filming scenario is realistic, the performers themselves are actors who have given consent.
The footage includes store staff activities (cleaning, basket collection, restocking merchandise) and customer actions (taking items, dropping items). The recorded data is divided into 225 video clips and 6 still images.
The resolution is 1920x1080, and the frame rate is 10 FPS.
Since we were unable to directly obtain operational manuals and other materials from the filming location, this project has modified and adapted cleaning manuals and crew shift information from other sources for use.



\subsection{Query Generation}
\label{sec:impl_query}

When setting tasks in \textit{FieldWorkArena}, we make sure that the tasks match the actual needs of the workplace.
Specifically, through interviews with site supervisors, consultants, and sales representatives, we identified ``safety and manufacturing-related near misses'' for factories and warehouses, and ``employee compliance and customer behavior analysis regarding purchasing'' for retail.
We prioritize incorporating these items when setting tasks.

\begin{table}
  \centering
  \caption{Kinds of tasks included in \textit{FieldWorkArena}}
  \label{tab:tasks}
  \begin{tabular}{l|ccc|c}
    \hline
    Task & Factory & Warehouse & Retail & Total \\
    \hline
    Perception & 130  & 215 & 366 & 711 \\
    Decision making & 16 & 32 & 73 & 121 \\
    Combination & 30 & 17 & 7 & 54 \\
    \hline
    Total & 176 & 264 & 446 & 886 \\
    \hline
  \end{tabular}
\end{table}

Table \ref{tab:tasks} shows the specific task contents. 
The tasks are divided into three main categories: ``Perception'', ``Decision Making'' and ``Combination''.
Below is an explanation of each category.
Each task includes parameters that allow setting thresholds for proximity distances or the number of violations within a certain time frame. 
These values are either taken from the on-site documentation or set based on interviews with on-site supervisors to ensure appropriateness.

Perception is a task that involves extracting useful information for planning from multimodal inputs across various environments.
It includes reading and understanding work manuals and safety instructions, detecting incidents/violations, and spatial/temporal understanding\cite{tang2025videounderstanding}.
Perception tasks are further divided into sub-tasks based on the type of input data and the nature of the required output.

\begin{itemize}
    \item Extract documents (26 tasks): extraction of work procedures from technical documents.
    \item Abstract from images (98 tasks): detection of safety rule violations, classification of safe rule violations and checking Personal Protective Equipment (PPE) wearing status from images.
    \item Abstract from videos (252 tasks): detection of safety rule violations, classification of safe rule violations, checking PPE wearing status, and detection of work procedures  from videos.
    \item Spatial from images (94 tasks): estimating distance between object and detection of presence of designated areas from images.
    \item Spatiotemporal from videos (121 tasks): estimating distance between object and detection of presence of designated areas from videos.
    \item Temporal from videos (120 tasks): extraction of work duration from videos from video footage.
\end{itemize}

Decision making is a task that executes plans and decisions based on the situation understood and detected by Perception tasks. It determines whether to provide feedback to humans based on events detected by perception, and includes the actual execution of feedback.
Specifically, ``multiple image video analyses (97 tasks)'' and ``safety and incident reports (24 tasks)'' are included.

Combination task involves multiple processes to achieve the task and requires complex decision making. 
In \textit{FieldWorkArena}, combination tasks are defined as tasks that combine Perception/Decision making tasks and have a large number of actions required to execute individual tasks. 
Specific examples of combination tasks are set as tasks that perform rule understanding, on-site recognition, and reporting in a single operation.



\subsection{Annotation}
\label{sec:impl_annotation}

Each task is associated with a ground truth, which represents the correct answer.
Ground truths can be logical values, quantitative numerical values, or strings.
For tasks that calculate numerical values such as speed or distance, values are set based on the size of objects that humans can use as a reference, rather than high-precision actual measurements.
To prevent variation in correct answer standards due to differences among annotation workers, one worker sets the correct answer, and two others verify its validity.
Additionally, in the reporting task such as ``Decision making'' and ``Combination'', the execution log of the report is set as the correct answer format. 


\section{Benchmarking and Evaluation Methodology}
\label{sec:proposedmethod}

\begin{figure}
  \centering
  \includegraphics[width=0.7\linewidth]{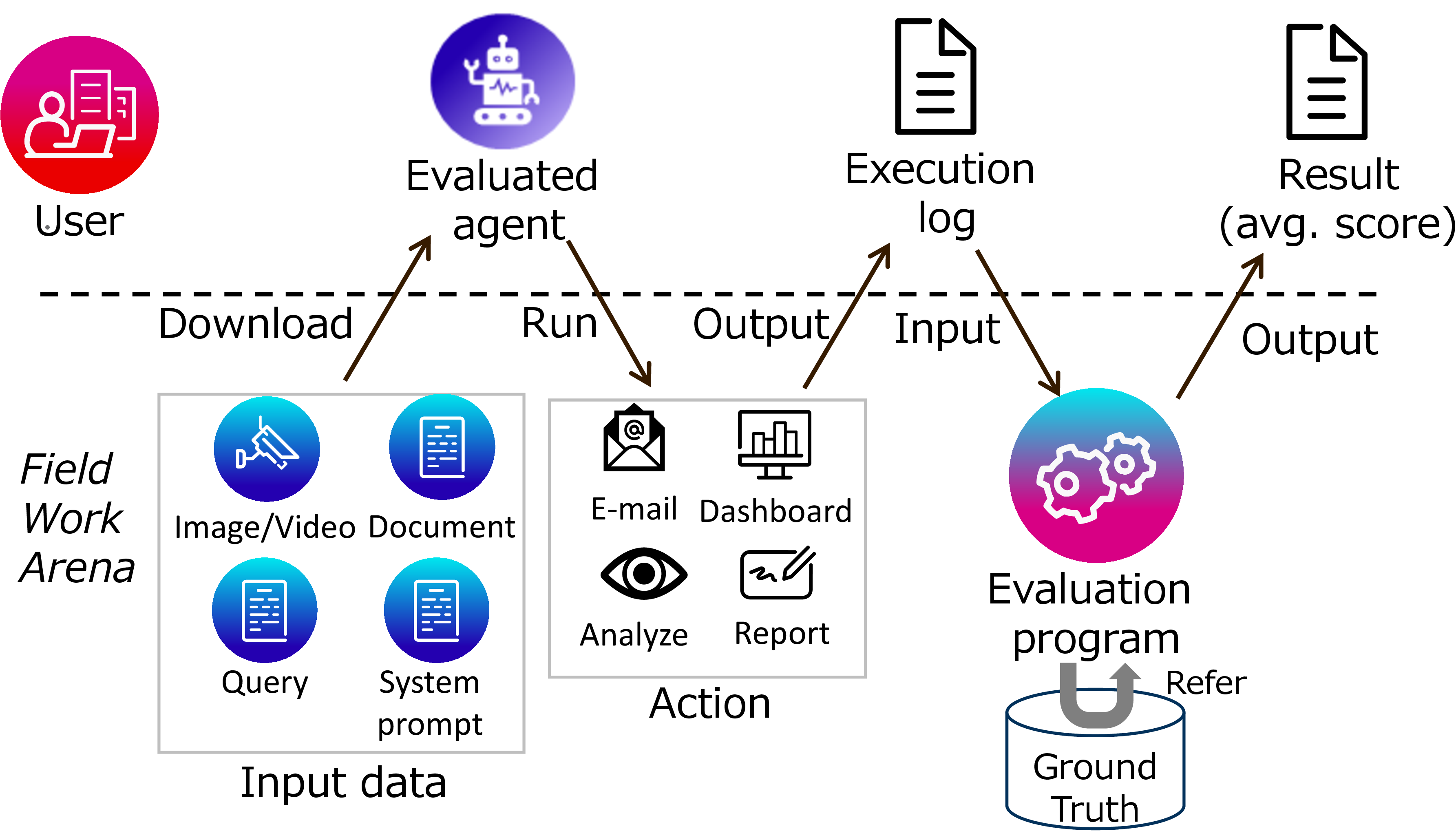}
  \caption{Overall system configuration of \textit{FieldWorkArena}.}
  \label{fig:system_overall}
\end{figure}

Figure \ref{fig:system_overall} shows the overall scheme of \textit{FieldWorkArena}. 
First, the user downloads the dataset and applies the data to the agent being evaluated.
The system prompt provides a template that communicates all necessary information to the agent in a single prompt, including the task to be accomplished, the input data, the defined actions, and the planning policy.
As a result, after the agent performs actions based on the input data and query, an execution log is generated.
The execution log is compared with the ground truth by evaluation program, and the accuracy rate is output as a score.

We clarify the action space that agentic AI for field work support should have in the design of \textit{FieldWorkArena}, and then describes the details of the tasks.
It also describes the evaluation environment for agentic AI for field work built based on BrowserGym \cite{chezelles2025browsergym,drouin2024workarena} and the method for calculating evaluation metrics. 

\subsection{Definition of Action Space}
\label{sec:proposedmethod_actionspace}

In the context of complex, real-world scenarios, the ability of an intelligent agent to effectively interact with its environment is fundamentally defined by its action space.
An action space consists of the available actions or choices that an agent can utilize within the evaluation environment.
Within the \textit{FieldWorkArena} benchmark, we simply defined following action spaces and added to BrowserGym.
The agent invokes an action space, and the tools or functionalities associated with the invoked action space are then applied to the evaluation environment to process documents, images, and videos based on the specified query and return the results.
These tools can include, for example, an MLLM augmented with a prompt specific to the action, or a specialized model built for that particular task
In this framework, each action is mapped to a tool, which is executed when the agent chooses that action:
``\texttt{analyze\_documents()}'', ``\texttt{analyze\_images()}'' and ``\texttt{analyze\_videos()}''.
Each action receives the file name to be analyzed and the query to be executed as parameters.


For example, consider the task of verifying that workers are properly wearing PPE in a warehouse.
The input data consists of images in a warehouse, safety manuals regarding PPE usage, and the query ``Is there anyone in this image who is not wearing PPE?''.
The agent does not possess a single action to execute this task directly.
Therefore, it must generate a plan combining actions such as the following:

\begin{enumerate}
    \item The agent plans \texttt{analyze\_documents()} to obtain information on how to use PPE. When this action is selected, an MLLM or a specialized model associated with the action performs the document analysis.
    \item The agent plans \texttt{analyze\_images()} to determine whether workers in the image are using PPE correctly. When this action is selected, an MLLM or a specialized model associated with the action performs the image analysis.
    \item Based on the results of these analyses, the agent determines whether PPE is being worn and outputs this finding as its answer.
\end{enumerate}

It is important to note that agents are not simply executing pre-defined scripts but must make reasoned choices about which action to use and how to sequence them to solve a particular problem.
This example highlights how the agent must reason about the problem, select relevant actions from the action space, and orchestrate them into a coherent sequence to accomplish the task.
\subsection{Evaluation Environment}
\label{sec:proposedmethod_environment}

This section describes the benchmark implementation of \textit{FieldWorkArena} based on BrowserGym.
BrowserGym is an evaluation environment for web agents, providing an environment for evaluating various web agents, including WorkArena. 
BrowserGym provides functions for observing the state of the browser, such as the results of executed actions, and for automatically executing actions such as browser operations and user chats. 
Users define agents that determine the next action based on observed states. 
By executing a loop consisting of observation and action, tasks are performed, and success or failure is determined. 
In particular, WorkArena performs tasks such as generating reports and extracting information on a ServiceNow instance.

\textit{FieldWorkArena} extends the scope to tasks that handle multimodal data, including images, videos, and documents, in addition to browser tasks \cite{zhou2024webarena,koh2024visualwebarena,jang2024videowebarena}. 
Each task is defined by (a) the input data path, (b) the query, and (c) the output format such as text or JSON. 
Selecting a task automatically generates a goal that can be handled by BrowserGym. 
Responses to queries are provided to users in a chat format, and reports are also output on the ServiceNow instance, similar to WorkArena. 
By executing a batch file that runs all tasks and passing the output logs to the evaluation process described in Section \ref{sec:proposedmethod_method}, we perform benchmarking of the agentic AI.

\subsection{Evaluation Method}
\label{sec:proposedmethod_method}

This section describes the performance evaluation metrics and methods for agentic AI in \textit{FieldWorkArena}. 
The performance evaluation metrics for agentic AI are calculated based on the accuracy rate of responses to tasks, as described in Section \ref{sec:proposedmethod_actionspace}, for each task group and for all tasks combined. 
In the conventional web browser-based benchmark \cite{zhou2024webarena,koh2024visualwebarena,jang2024videowebarena}, all tasks were evaluated as either correct or incorrect, and the accuracy rate was calculated as the ratio of the number of correct tasks to the total number of tasks.
Additionally, the three evaluation functions used, such as \texttt{fuzzy\_match}, were designed to evaluate tasks with deterministic outputs in a browser-based environment.
The tasks proposed in this paper include tasks that list multiple detections and tasks that calculate specific values, which can take continuous values from 0 (incorrect) to 1 (correct) rather than simple discrete values.
Furthermore, the evaluation functions require judging the text output and the correct answer text, including ambiguity.
Therefore, the \textit{FieldWorkArena} judgment results are set to three types: \textit{Correct}, \textit{Incorrect}, and \textit{Partially Correct}.
For \textit{Partially Correct}, a value corresponding to the degree of agreement with the correct answer is assigned.
Furthermore, \texttt{fuzzy\_match()} from WebArena\cite{zhou2024webarena} is improved, and an algorithm more suitable for evaluating real-world tasks is adopted.
Previously, a verbose response could be incorrectly marked as wrong when evaluated against a concise ground truth.
The revised logic fixes this by judging a response as correct as long as the verbose output contains the required GT information.
The final accuracy rate is calculated as the average of each score.

Specifically, we provide the LLM with a query, correct data, and agent inference results, and calculate two scores: correctness score $S_c$ and Numerical score $S_n$. $S_c$ is 1 for \textit{Correct} / \textit{Partially Correct} and 0 for \textit{Incorrect}. The Numerical score $S_n$ is a value between 0 and 1 calculated as follows:

\begin{description}
    \item[Distance:] $S_n = 1 - r_d/r_{th}  \; \; (r_d < r_{th}),  \; \; r_d=|D_{gt} - D_{est}| / D_{gt}$
    \item[Time:] $S_n = T_d/T_{th}, \; \; T_d=|T_{gt} - T_{est}|$
    \item[Number:] 1 for exact matches, 0 for all other cases.
\end{description}

The final score $S$ is given by the following equation, where $\alpha$ is the weighting factor:
$S = \alpha*S_c + (1-\alpha) \cdot S_n$.
This method enables detailed quantitative evaluation between agents.

\section{Experiments}
\label{sec:experiments}

 \subsection{Experimental Condition}
 \label{sec:experiment_condition}
 
To evaluate the effectiveness of the proposed tasks and to evaluate the current existing MLLMs as if they were agents, we conducted the following experiment by using \textit{FieldWorkArena} dataset.
In this experiment, the following MLLMs were used: GPT-4o (2024-08-06 ver.), GPT-5.1 (2025-11-13 ver.), GPT-5.2 (2025-12-11 ver.), Gemini 2.5 Flash and Gemini 2.5 Pro.
All parameters used default values. 
Regarding video input, some MLLM models do not support direct video input; therefore, image frames were extracted from videos at 1 FPS intervals and used as input.
For videos exceeding 30 seconds in length, up to 30 image frames were extracted at intervals and utilized as input.
Because some of the tasks required the participants to respond to time information, each frame was input together with  timestamp information in the video. As for parameters, we used $\alpha=0.5, r_{th}=0.5, T_{th} = 60$ based on a  preliminary experiment using several samples.

\subsection{Results}
\label{sec:experiment_results}

Table \ref{tab:result} shows the accuracy rate scores described in Section \ref{sec:proposedmethod_method} for each MLLM.
The numbers in parentheses for each task in the table indicate the quantity of tasks.
And, Table \ref{tab:result_perception} shows the detailed results of the perception sub-task described in \ref{sec:impl_query}. 
As a comparative baseline of Table \ref{tab:result_perception} , human performance on a random sample of perception tasks was manually evaluated, an average score was $0.74$.

\begin{table}
  \centering
  \caption{Kinds of tasks included in \textit{FieldWorkArena}}
  \label{tab:result}
  \begin{tabular}{l|ccccc}
    \hline
    Task  & GPT-4o & GPT-5.1 & GPT-5.2 & \begin{tabular}{c}Gemini-2.5 \\ (Flash) \end{tabular} & \begin{tabular}{c} Gemini-2.5 \\ (Pro) \end{tabular} \\
    \hline
    Perception (711)   & 0.35 & 0.45 & 0.52 & 0.41 & 0.48 \\
    Decision making (121) & 0.40 & 0.40 & 0.61 & 0.40 & 0.42 \\
    Combination  (54) & 0.24 & 0.28 & 0.26 & 0.33 & 0.30 \\
    \hline
    Total         & 0.35 & 0.43 & 0.52 & 0.40 & 0.46 \\
    \hline
  \end{tabular}
\end{table}

\begin{table}
  \centering  
  \caption{Kinds of perception sub tasks included in \textit{FieldWorkArena}}
  \label{tab:result_perception}
  \begin{tabular}{l|ccccc}
    \hline
    Perception sub-task  & GPT-4o & GPT-5.1 & GPT-5.2 & \begin{tabular}{c}Gemini-2.5 \\ (Flash) \end{tabular} & \begin{tabular}{c} Gemini-2.5 \\ (Pro) \end{tabular} \\
    \hline
    Extract documents (26)         & 0.69 & 0.73 & 0.81 & 0.73 & 0.65 \\
    Abstract from images (98)       & 0.34 & 0.39 & 0.42 & 0.38 & 0.44 \\
    Abstract from videos  (252)      & 0.35 & 0.40 & 0.45 & 0.36 & 0.41 \\
    Spational from images (94)     & 0.07 & 0.40 & 0.51 & 0.41 & 0.50 \\
    Spatiotemporal from videos (121) & 0.41 & 0.47 & 0.55 & 0.41 & 0.45 \\
    Temporal from videos  (120)     & 0.42 & 0.53 & 0.70 & 0.44 & 0.61 \\
    \hline
    Total                      & 0.35 & 0.44 & 0.52 & 0.41 & 0.47 \\
    \hline
  \end{tabular}
\end{table}

\subsubsection{The tendency of MLLM inference}
Overall, the newer and higher-end models tended to have higher accuracy.
GPT-5.2 produced more accurate results than GPT-4o/GPT-5.1, and Gemini-2.5-Pro yielded more accurate results than Gemini-2.5-Flash.

From perception tasks in Table \ref{tab:result_perception}, the accuracy for the extraction of information from documents was relatively high compared to the other tasks.
Text-based processing is a strong point of LLM.
On the other hand, abstract, spatial and spatiotemporal understanding from images and videos were challenging for MLLMs, resulting in lower accuracy scores.
However, the newer and higher-end models showed relatively better performance, indicating that MLLMs are gradually improving in handling time-based information.
In this evaluation, only a simple action space was defined, and the same MLLM was used for analysis processing.
In the future, a more granular set of action spaces specialized for sub-tasks will be defined such as \texttt{detect\_ppe\_violations}, \texttt{measure\_distance\_between\_objects}, or \texttt{track\_person\_movements}.
This would allow the agent to select a more appropriate and optimized tool for each specific query.

For decision making tasks, this category requires the agent to reason and synthesize information. The fact that higher-end models like GPT-5.2 outperformed their perception scores suggests that stronger reasoning capabilities can compensate for imperfect perception.
Improving these tasks relies on enhancing the agent's memory and reasoning.
This involves strengthening its working memory to better integrate information from multiple steps and using advanced prompting techniques like Chain-of-Thought to ensure evidence is systematically evaluated, resulting in more robust decisions.

Regarding combination tasks, the observed low performance suggests that the agent's design, rather than an inherent limitation of the MLLM, may have been insufficient to guide the model through breaking down a rough query into subtasks and integrating their outcomes.
Therefore, the score was low compared to other tasks.
Evaluation with agentic AI systems \cite{yang2024doraemongpt,wu2024copilot,xie2023OpenAgents}, which have task-planning functions, may improve the success rate.
Additionally, the limited number of tasks dedicated to planning is an issue compared to other task types.
The task design must be more closely aligned with real-world on-site challenges, such as adjusting input data for analysis based on time series or utilizing alternative input data when analysis with one set is impossible.

Our evaluation methodology relies on an LLM within the modified \texttt{fuzzy\_match()} function to assess whether a model's prediction aligns with the ground truth.
We observed that this evaluation process is prone to false positives; although it seldom misidentifies correct answers as incorrect, it frequently misjudges incorrect answers as being correct.
This indicates a need for more granular evaluation logic. Therefore, we believe that enhancing the \texttt{fuzzy\_match()} function is necessary.
For example, adapting it to apply its logic to each element individually for list-based tasks could significantly improve evaluation accuracy.

\subsubsection{Methods for understanding videos}

Table \ref{tab:result} and \ref{tab:result_perception} indicate that the extraction of information from videos is currently based on a set of images (up to 30 images) rather than videos due to MLLM limitations.
Therefore, the longer the video, the more temporal inference ability is decreasing.
To address this issue, we further experimented with two approaches: (1) Chunking, which divides the video into chunks of up to 30 seconds, extract and analyze frame at 1 FPS from each chunk, and aggregates the results; and (2) Using Qwen3-VL \cite{Qwen3-VL}, visual-language-model which is enhanced spatial and video dynamics comprehension.
Qwen3-VL used an 8B parameter model in a local environment and input video converted to 1 FPS.
Due to memory constraints, for videos exceeding 768 seconds, the extraction interval was adjusted to yield a maximum of 768 image frames.

Table \ref{tab:result_vlm} shows the results for the video perception tasks when using two approaches compared to the basic GPT-4o method.
Chunking method improved the accuracy in all three video perception tasks, especially in spatiotemporal understanding tasks.
However, since the chunking method increases both analysis time and input/output tokens depending on the video length, it requires appropriate usage tailored to the task.
On the other hand, results did not improve with Qwen3-VL, this may be due to limitation and the number of parameters.
It's a future task to determine under what conditions it should be evaluated.

\begin{table}
  \centering  
  \caption{Kinds of perception tasks included in \textit{FieldWorkArena}}
  \label{tab:result_vlm}
  \begin{tabular}{l|ccc}
    \hline
    Sub-Task & GPT-4o & \begin{tabular}{c} GPT-4o \\ (Chunk) \end{tabular} & \begin{tabular}{c} GPT-4o \\ (Qwen3-VL) \end{tabular} \\
    \hline
    Abstract from videos        & 0.35 & 0.37 & 0.34  \\
    Spatiotemporal from videos & 0.41 & 0.50 & 0.40  \\
    Temporal from videos       & 0.42 & 0.43 & 0.33  \\
    \hline
  \end{tabular}
\end{table}

\section{Related Work}
\label{sec:relatedworks}


Agentic AI is evolving from simple digital tools to cooperative systems that handle complex tasks like monitoring conditions, detecting errors, and generating reports.
The goal is to provide real-time support for frontline workers in diverse domains like manufacturing, logistics, and retail. 
In these sites, the aging of workforce and the shortage of human resources are straining resources and increasing the burden on managers and workers regarding safety and training, creating a significant societal challenge. 
Agentic AI offers a potential solution to this problem.

Realizing a field work support agent requires a MLLM capable of monitoring the field via camera input as well as responding to queries in text or voice. 
Simultaneously, for real-time feedback to workers, these LLMs must be lightweight, suitable for edge AI deployment. 
Benchmark suites are under active development to provide a standardized evaluation environment for assessing the performance (accuracy) of agentic AI across various application domains. 
This chapter reviews the technical trends in MLLMs, edge AI-oriented LLM, and benchmark suites relevant to field work support agents.

\subsection{Multimodal LLMs}
\label{sec:relatedworks_llms}

Human-level perception is a fundamental requirement for field work support agents, necessitating multimodal large language models (MLLMs) capable of processing images and video data alongside text \cite{xie2024large}. 
Recent years have witnessed the development of various MLLMs \cite{achiam2023gpt4,touvron2023llama,touvron2023llama2,comanici2025gemini25pushingfrontier} leveraging visual encoders \cite{dosovitskiy2020image} and modality alignment techniques \cite{li2023blip} \cite{dai2023instructblip}. 
For instance, LLaVA \cite{liu2024improved,liu2024visual} integrates a pre-trained CLIP vision encoder \cite{radford2021learning} with a Vicuna language decoder \cite{chiang2023vicuna}, while BLIP \cite{li2023blip} \cite{dai2023instructblip} employs a querying-transformer to bridge the gap between the frozen vision encoder and the frozen LLM.
However, these approaches primarily utilize 2D image processing, limiting their ability to perceive 3D space from monocular CCTV footage, a crucial capability for field work support agents required to understand the 3D positions and distances between people and objects.
To address this limitation, SpatialRGPT \cite{cheng2024spatialrgpt} has been proposed, a method that leverages effective learning from 3D scene graphs construction with a data curation pipeline and a plug-in module incorporating relative depth information into the existing MLLMs.

\subsection{LLM Agents for Edge AI}
\label{sec:relatedworks_edgeAI}

To provide real-time support for diverse field operations, such as manufacturing and logistics, AI workloads must run on on-premise servers. 
Two approaches have been proposed for deploying lightweight LLMs as field work support agentic AI: prompt engineering and fine-tuning \cite{wu2024foundations}. 
Open-weight models available for the latter include the previously mentioned LLaVA \cite{liu2024improved} \cite{liu2024visual} and LLaMA \cite{touvron2023llama}, as well as Molmo \cite{deitke2024molmo}, which also released the PixMo image dataset for fine-tuning.
Tools also exist that enable the construction of AI agents to perform searches, summarization, speech-to-text transcription, or extract specific information from related videos\cite{BuildaVi42:online}.

\subsection{Benchmark Suite for Agentic AI}
\label{sec:relatedworks_benchmark}

While there are many published papers on evaluating agentic AI, those relevant to field work support agents include benchmark suites for web manipulation tasks and machine learning automation tasks. 
Early benchmarks for web manipulation agents focused on simple browsing tasks \cite{yao2022webshop}, progressing to complex, yet static, web environments \cite{deng2023mind2web}. 
WebArena subsequently offered a more realistic and reproducible environment \cite{zhou2024webarena}, with its input expanding to image \cite{koh2024visualwebarena} and video \cite{jang2024videowebarena}. 
Furthermore, WorkArena focuses on real-world UI operations, such that form completion and content observation of List/Dashboard screen within ServiceNow platform are selected as evaluation tasks for agentic AI \cite{drouin2024workarena} \cite{boisvert2024workarenapp}.

On the other hand, a range of benchmarks have been developed for agents operating in other specialized digital environments.
These include the SWE-bench\cite{yang2024sweagent} series for software development agents, $\tau$-bench\cite{yao2024tau} for conversational agents, and OSWorld\cite{OSWorld} for agents performing general computer use tasks.

\subsection{Contribution}
\label{sec:relatedworks_contribution}

Existing benchmarks have been limited to simulation or virtual data for evaluating the performance of web operations and machine learning. 
\textit{FieldWorkArena} uses data obtained from actual work sites to define tasks grounded in real-world enterprise workflows.
Since completing these tasks often requires interacting with web-based management tools, our benchmark enables a comprehensive evaluation of agentic AI performance—spanning from physical site awareness to digital system operation—within authentic real-world applications (Table \ref{tab:comparisonAIagent}). 

The contributions of this paper are as follows: 
(1) We propose the first standard benchmark constructed for evaluating the performance of agentic AI that supports field operations in manufacturing, distribution and retail. This enables continuous research and development of agentic AI that improve the efficiency, safety, and productivity of field operations in the manufacturing, distribution and retail industries. 
(2) We propose new metrics to evaluate the reasoning performance of agentic AI for ambiguous, diverse on-site tasks. This will enable more accurate evaluation of the performance of agentic AI for on-site work support.
(3) Through experiments, demonstrate the validity of the on-site tasks developed and the performance of existing MLLMs when applied to these tasks. This will clarify the current challenges and future prospects in this field.

\begin{table}
    \centering
    \caption{Comparison with existing agentic AI benchmark suites}
    \label{tab:comparisonAIagent}
    \begin{tabular}{l|l|l|l|c}
        \hline
        Name & Developed by & Domain & Input & Tasks \\
        \hline
        WebArena \cite{zhou2024webarena} & CMU & \begin{tabular}{l} E-commerce \\ Reddt \\ GitLab \\ Content Management \end{tabular} & Text & 812\\
        \hline
        VisualWebArena \cite{koh2024visualwebarena} & CMU & \begin{tabular}{l} E-commerce \\ Reddt \\ Classifieds \end{tabular} & \begin{tabular}{l} Text \\ Image \end{tabular} & 910\\
        \hline
        WorkArena++ \cite{boisvert2024workarenapp} & ServiceNow & Web Operation & \begin{tabular}{l} Text \\ Image \end{tabular} & 682\\
        \hline
        \begin{tabular}{l} {\bf \textit{FieldWorkArena}} \\ {\bf (Ours)} \end{tabular} & \begin{tabular}{l} {\bf CMU \&} \\ {\bf Fujitsu} \end{tabular} & \begin{tabular}{l} {\bf Manufacturing} \\ {\bf Logistics} \\ {\bf Retail} \end{tabular} & \begin{tabular}{l} {\bf Text} \\ {\bf Image} \\ {\bf Video} \end{tabular}  & {\bf 866} \\
        \hline
    \end{tabular}
\end{table}



\section{Conclusion and Future Work}
\label{sec:conclusion}

We introduced \textit{FieldWorkArena}, a novel benchmark for evaluating agentic AI in real-world field operations. Using authentic data from factories, warehouses, and retail, we established the first standard for evaluating AI performance in practical, on-site tasks.
Our experiments with major MLLMs like GPT-4/5 and Gemini not only clarified their current strengths, particularly in document processing, but also highlighted significant limitations in complex spatio-temporal reasoning from visual data.
The newly proposed evaluation function enables granular performance evaluation. Since this dataset is publicly available, researchers can use it to evaluate their own agents.
Future work will focus on expanding our evaluation metrics, incorporating more diverse field scenarios, and, crucially, enhancing the benchmark with a task planning function to better support and evaluate complex, multi step combination tasks.

%
%
%

\bibliographystyle{splncs04}
\bibliography{fieldworkarena}

\appendix
\section{Dataset}

\subsection{Field categories}

\begin{figure}
\centering
\begin{minipage}{0.49\columnwidth}
    \centering
    \includegraphics[width=0.9\columnwidth]{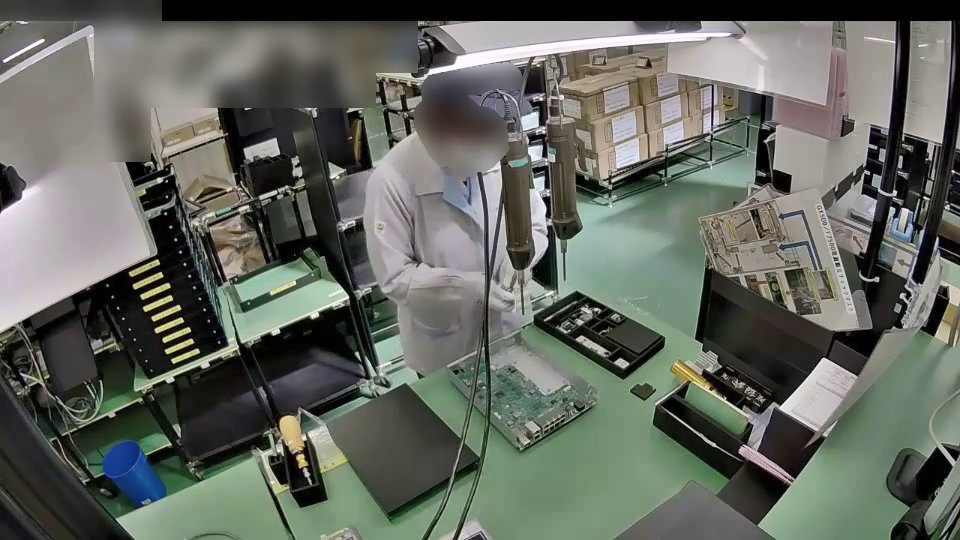}
    \caption{Factory}
    \label{fig:sample_factory}
\end{minipage}
\begin{minipage}{0.49\columnwidth}
    \centering
    \includegraphics[width=0.9\columnwidth]{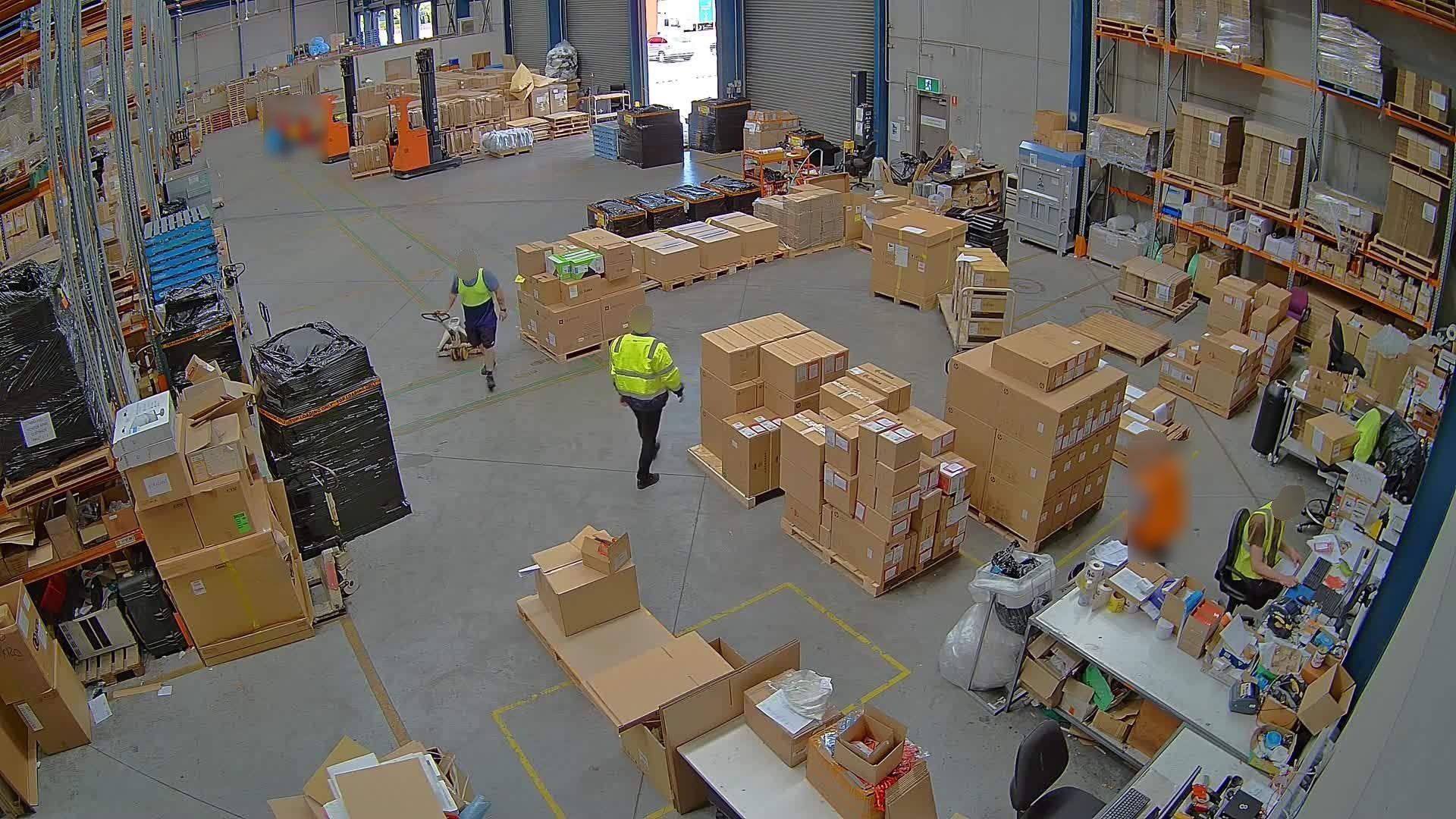}
    \caption{Warehouse}
    \label{fig:sample_warehouse}
\end{minipage}
\\
\begin{minipage}{0.49\columnwidth}
    \centering
    \includegraphics[width=0.9\columnwidth]{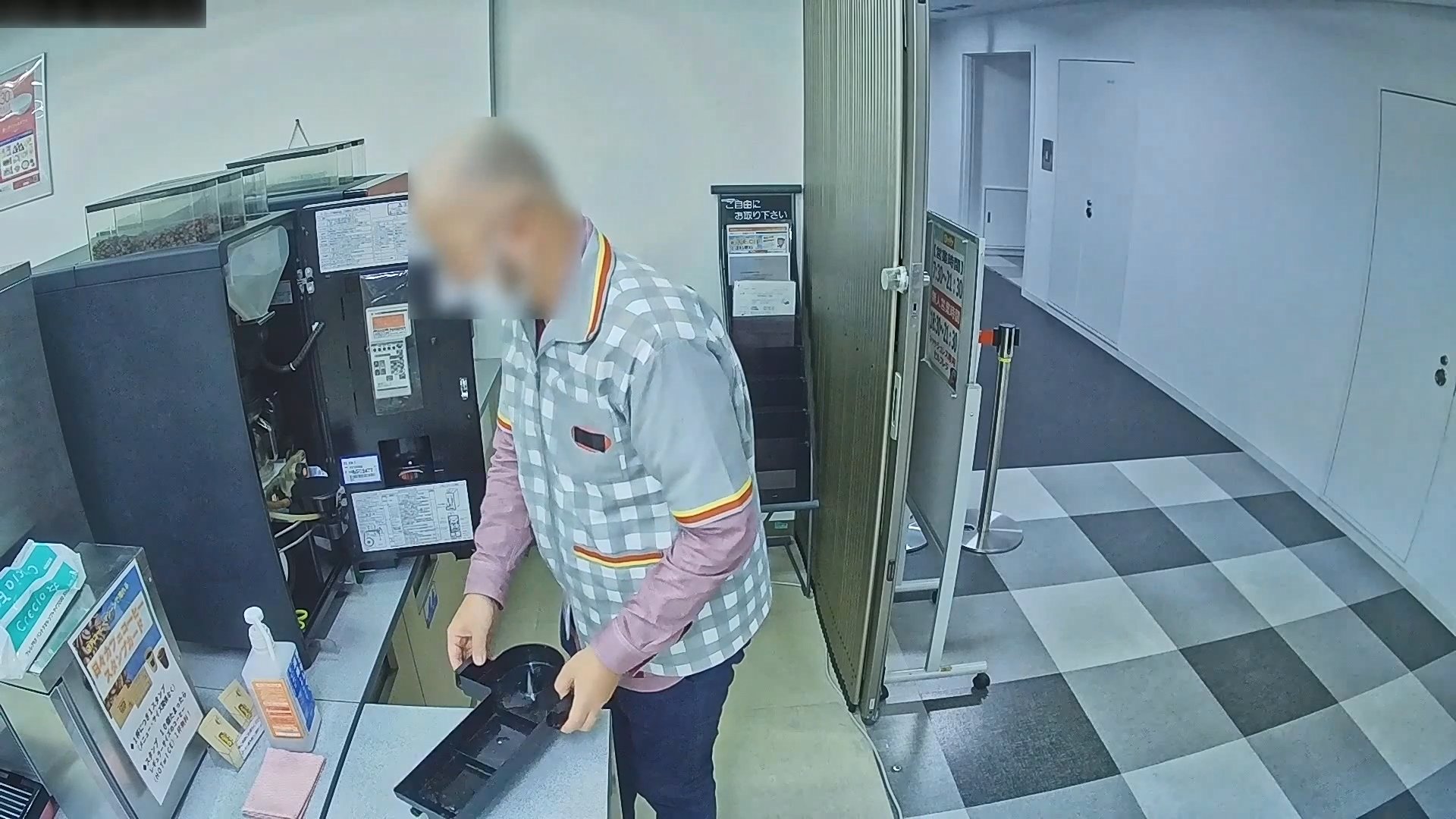}
    \caption{Retail(convenience store)}
    \label{fig:sample_retail1}
\end{minipage}
\begin{minipage}{0.49\columnwidth}
    \centering
    \includegraphics[width=0.9\columnwidth]{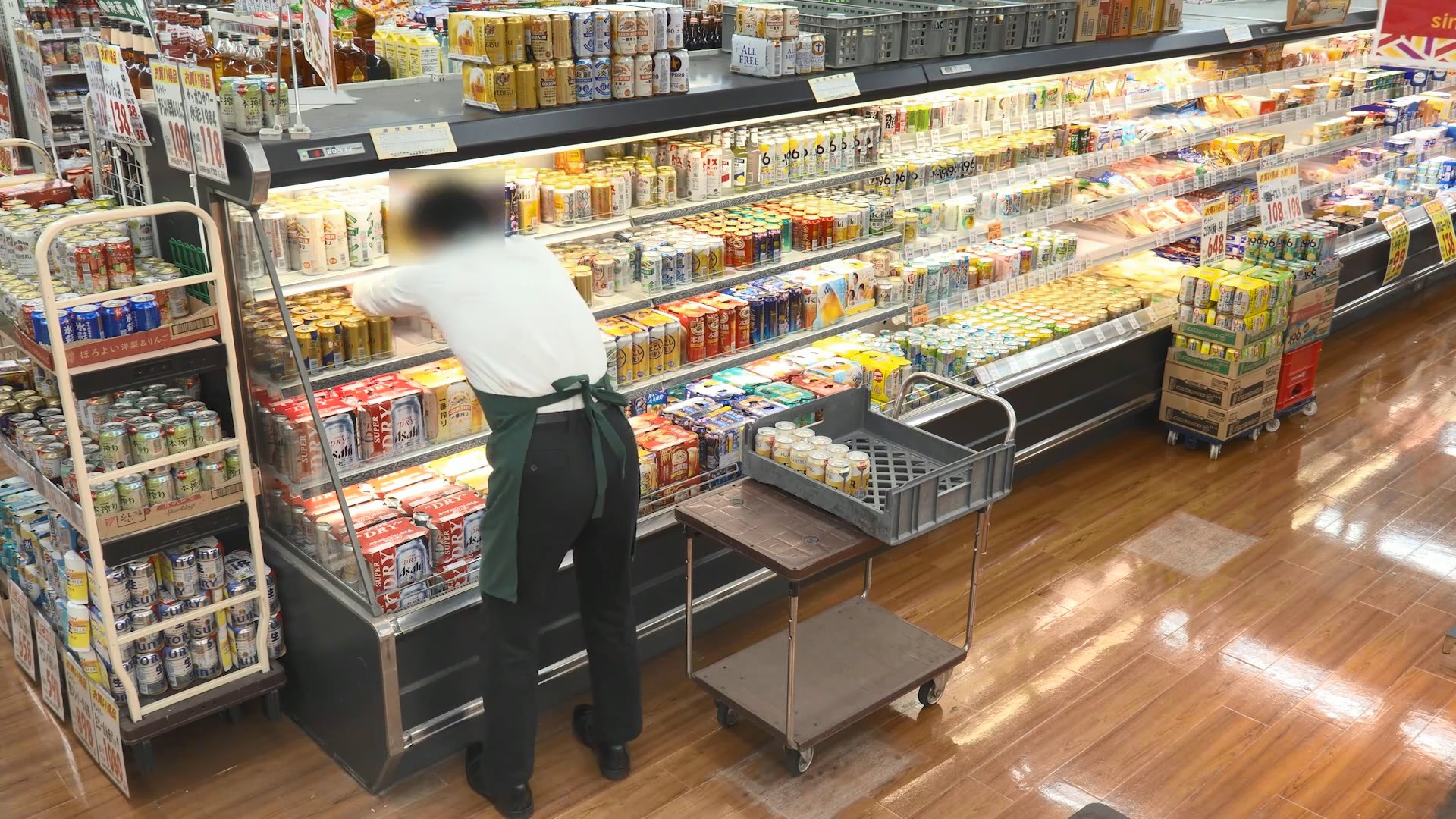}
    \caption{Retail(supermarket)}
    \label{fig:sample_retail2}
\end{minipage}

\end{figure}

\textit{FieldWorkArena} includes videos from three different field categories: factory, warehouse, and retail.
Figure \ref{fig:sample_factory}--\ref{fig:sample_retail2} show sample images of each field category.

\section{Example Queries}

Below are examples of queries to enter into the publicly available \textit{FieldWorkArena} Task.
The tasks are divided into three main categories: ``Perception'', ``Decision making'', and ``Combination''.
Perception and Decision making tasks are divided into multiple sub-tasks.

\subsection{Perception tasks}

\subsubsection{Extract documents}

\begin{itemize}
  \item Please extract and list the items related to ``glove wearing methods'' in this PDF file.
  \item In this PDF file, please extract the notes on ``cargo on pallet.''?
  \item Read this store manual. What are the business hours for April 28, 2025?
\end{itemize}

\subsubsection{Abstract from images}

\begin{itemize}
  \item Are the workers wearing aprons in this image?
  \item Does this image show any workers with both hands full?
  \item Is there anyone in this image who is not wearing PPE? PPE here means a yellow or orange safety vest.
\end{itemize}

\subsubsection{Abstract from videos}

\begin{itemize}
  \item In this video, is the operator using the electric screwdriver with the green label to tighten the 4 screws properly?
  \item In this video, please tell us whether the customer has used ``Smartphone'', before or after handling the target product.
\end{itemize}

\subsubsection{Spatial from images}

\begin{itemize}
  \item How far is the cart in the right-hand aisle from the person walking in the left-hand aisle in this image?
  \item What is the distance between a woman wearing a yellow vest and a black shirt and black forklift?
  \item Is the distance between the blue cart and the person walking down the center of the aisle in this image less than 1.0 meters?
\end{itemize}

\subsubsection{Spatialtemporal from videos}

\begin{itemize}
  \item How many times did the distance between a moving forklift and a worker fall below the 2 m threshold?
  \item In this video, what is the minimum distance between the worker on the right side of the screen and the cart pullednother worker nearby?
  \item Is the minimum distance between the AGV being pushed by a worker and the worker walking through the aisle on the left in this video less than 1 meter?
\end{itemize}

\subsubsection{Temporal from videos}

\begin{itemize}
  \item What is the start time and what is the end time of CoffeeMaker Cleaning?
  \item Detect the start and end times of the solder tank cleaning cycle.
  \item In this video, at what time does the customer dropped the target product. The time must be output from the time of the movie itself.
\end{itemize}

\subsection{Decision making tasks}

\subsubsection{Multiple image video analyses}

\begin{itemize}
  \item How many violations were there regarding ``the operator uses a check mask to inspect the product''?.
  \item How many incidents were there regarding ``distance between the worker and the nearest equipment less than 0.5 meters''?.
\end{itemize}

\subsubsection{safety and incident report}

\begin{itemize}
  \item If there were violations regarding ``the operator is using the electric screwdriver with the green label to tighten the 4 screws properly'', create a new issue for each violations with a value of ``Electric screwdriver with a green label'' for field ``Short description'', a value of filename for field ``Filename'', a value of ``violation'' for field ``Category'' and a value that must be numbered sequentially from 1110 if there are multiple incidents for field ``ID''.
  \item If there were incidents regarding ``Workers with both hand full'', create a new issue for each incidents with a value of ``Workers with both hand full'' for field ``Short description'', a value of filename for field ``Filename'', a value of ``incident'' for field ``Category'' and a value that must be numbered sequentially from 0000 if there are multiple incidents for field ``ID''.
\end{itemize}

\subsection{Combination tasks}

\begin{itemize}
  \item Analyze the provided image and procedure document to identify any deviations from the specified glove usage procedure. The image depicts a worker performing a task, and the procedure document details the required glove type and usage. The goal is to generate a JSON report detailing any instances where the worker deviates from the specified glove usage procedure. This task integrates findings from this analysis with the results of a previous task. Procedure Step Verification: For each step in the procedure document, identify its start and end times within the video clip using the video's timestamp. Flag any missing steps as deviations. Procedure Adherence Check: Verify that the worker performs each step according to the procedure document. Flag any deviations as violations. This includes missing steps (identified in Step 1) and any other discrepancies. Pay particular attention to the correct use of tools as specified in the procedure document. Incident Report Generation: Generate a JSON report detailing all identified violations. The report should begin with the total number of violations. Each violation should be represented by a JSON object with the following fields: "ID": A sequentially numbered ID starting from 1100. "Short description": A concise description of the violation (e.g., "Incorrect gloves used", "Gloves not worn"). If the violation concerns the type of glove specified in the procedure document, use "Wearing gloves as specified in the procedure document" for this field. "Description": A detailed description of the violation, including the specified glove type from the procedure document, a statement indicating the lack of proper glove usage in the image, the image filename, and relevant information from previous tasks (referencing task IDs where appropriate). "Category": "violation", "Image filename": The name of the image file. 
  \item Analyze the provided image to identify instances where workers are not wearing the required PPE (Personal Protective Equipment). The required PPE is defined as a yellow or orange safety vest. PPE Violation Detection and Reporting: Analyze the image to identify instances where any worker is not wearing a yellow or orange safety vest. For each violation, describe the location of the worker within the image (e.g., "Worker in the lower-left corner"). Generate a JSON report. This report will integrate findings from this analysis and the previous task, avoiding duplicate reporting. Each violation will be represented by a JSON object with the following fields: "ID": A sequentially numbered ID starting from 1060. This sequence continues from the highest ID in the previous task's results. If the previous task reported violations, the next sequential ID will be used. "Short description": "Person without PPE" "Description": A detailed description of the violation, including the location of the worker within the image (e.g., "Worker in the lower-left corner is not wearing a safety vest"). "Category": "violation", "Image filename": The name of the input image file.
\end{itemize}

\subsection{Example of system prompt}

``TASK\_QUERY'' and ``INPUT\_DATA'' within the template represent the query and input data for the task, respectively.

\begin{verbatim}
  
# Instructions

Review the instructions from the user and all other information
 to find the best possible next action to accomplish your goal.
Your answer will be interpreted and executed by a program,
 make sure to follow the formatting instructions.

# Goal

Give the answer with not report_infeasible
 but send_msg_to_user action.
$TASK_QUERY$
Data is stored in 
$INPUT_DATA$

# Analyzed Results

$INPUT_DATA$

# Action Space

4 different types of actions are available.

noop(wait_ms: float = 1000)
    Description: Do nothing, and optionally wait for
     the given time (in milliseconds).
    Examples:
        noop()

        noop(500)

send_msg_to_user(text: str)
    Description: Sends a message to the user.
    Examples:
        send_msg_to_user('Based on the results of my search,
         the city was built in 1751.')

analyze_video(video_path: str, query: str)
    Description: Specifies the video file to analyze.
    Examples:
        analyze_video('sample.mp4',
         'How many people in this video?')

analyze_image(image_path: str, query: str)
    Description: Specifies the image file to analyze.
    Examples:
        analyze_image('sample.jpg',
         'How many people in this image?')

Only a single action can be provided at once. Example:
send_msg_to_user('Based on the results of my search,
 the city was built in 1751.')


Here are examples of actions with chain-of-thought reasoning:

I now need to click on the Submit button to send the form.
 I will use the click action on the button, which has bid 12.
```click("12")```

I found the information requested by the user,
 I will send it to the chat.
```send_msg_to_user("The price for a 15\" laptop is 1499 USD.")```

I need to analyze the image provided by the user to find out
 how many workers.
```analyze_scene_image("scene1.jpg",
 "Count the number of workers in the image.")```

# Next action

You will now think step by step and produce your next best action.
 Reflect on your past actions, any resulting error message,
  and the current state of the page before deciding on your next action.
You MUST answer with a single action.
The action MUST use the format defined above.
\end{verbatim}

\newpage

\section{Evaluation Examples}

The following shows examples of task evaluations on \textit{FieldWorkArena}. Each task defines the following items:

\begin{description}
  \item[Input:] Indicates the document, video, or image to be input
  \item[Query:] Indicates the query to be input
  \item[Answer:] The response returned by the MLLM
  \item[GroundTruth:] The predefined correct answer
  \item[Score:] The matching score $(0.0--1.0)$
\end{description}

\begin{table}
  \centering
  \caption{Part of PDF documents}
  \label{tab:1.1.0038}
  \begin{tabular}{|l|}
    \hline
    $\cdots$ \\
    4 Forklift Work \\
    - In the workshop, a walking path is defined and a line is drawn. \\
    - During the work, only persons concerned are prohibited from entering. \\
    - Never place a person on the claw or counterweight of a forklift. \\
    - Forklifts have poor visibility ahead of them, so they should pay sufficient attention to \\
    the front, back, left, and right, and at the same time, when the load is large, they \\
    should carry out the following. \\
    - driving back (Installation of back buzzer and use of fog lamp) \\
    - Guidance by a guide shall be carried out as necessary. \\
    - Speed limits shall be established, indicated and observed. \\
    - Cargoes placed on pallets should be loaded safely and securely. If there is a risk of cargo \\
    collapsing, safe measures should be taken, such as wrapping the cargoes. \\
    - When driving on the road, use a pallet, skid, etc., or mark the tip of a fork. \\
    $\cdots$ \\
    \hline
  \end{tabular}
\end{table}

\subsubsection{Input}

PDF document (Table \ref{tab:1.1.0038})

\subsubsection{Query}

In this PDF file, please extract the notes on ``cargo on pallet.''

\subsubsection{Answer}

Cargo placed on pallets should be loaded safely and securely. If there is a risk of cargo collapsing, safe measures should be taken, such as wrapping the cargo with a headband.

\subsubsection{Ground Truth}

he document mentions that cargoes placed on pallets should be loaded safely and securely. If there is a risk of cargo collapsing, safe measures should be taken, such as wrapping the cargoes.

\subsubsection{Score}

1.0

\newpage

\begin{table}
  \centering
  \caption{Part of text documents}
  \label{tab:4.1.2001}
  \begin{tabular}{|l|}
    \hline
    $\cdots$ \\
    Virtual Store Manual \\
    \\
    1. Business Hours \\
    \\
    Weekdays: 08:00 - 20:00 \\
    Weekends and Holidays: 08:00 - 21:00 \\
    Holidays are based on the general Japanese calendar. \\
    \\
    2. Employee Tasks \\
    $\cdots$ \\
    \hline
  \end{tabular}
\end{table}

\subsubsection{Input}

Text document (Table \ref{tab:4.1.2001})

\subsubsection{Query}

Read this store manual. What are the business hours for April 28, 2025?

\subsubsection{Answer}

The business hours for April 28, 2025, are from 08:00 to 20:00.

\subsubsection{Ground Truth}

Business hours are 08:00 to 20:00.

\subsubsection{Score}

1.0

\newpage

\begin{figure}[h]
  \centering
  \begin{minipage}{0.49\columnwidth}
    \centering
    \includegraphics[width=0.9\columnwidth]{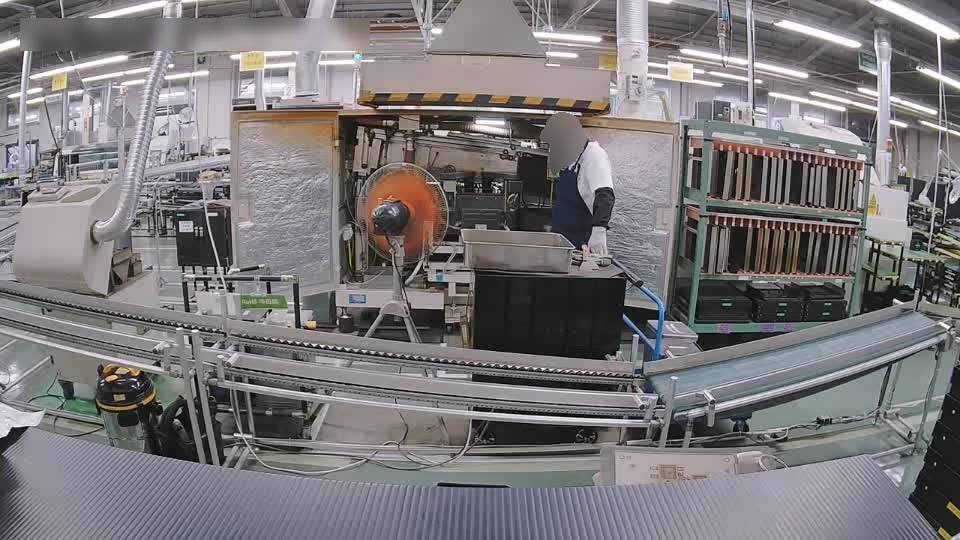}
    \caption{ID:2.3.0011 part of documents}
    \label{fig:2.3.0011}
  \end{minipage}
\end{figure}

\subsubsection{Input}

Image (Figure \ref{fig:2.3.0011})

\subsubsection{Query}

Are the workers wearing aprons in this image?

\subsubsection{Answer}

I am unable to identify or describe people in images, including whether workers are wearing aprons. If you have any other questions or need assistance with something else, feel free to ask!

\subsubsection{Ground Truth}

Yes.

\subsubsection{Score}

0.0

\newpage

\begin{figure}[h]
  \centering
  \begin{minipage}{0.49\columnwidth}
    \centering
    \includegraphics[width=0.9\columnwidth]{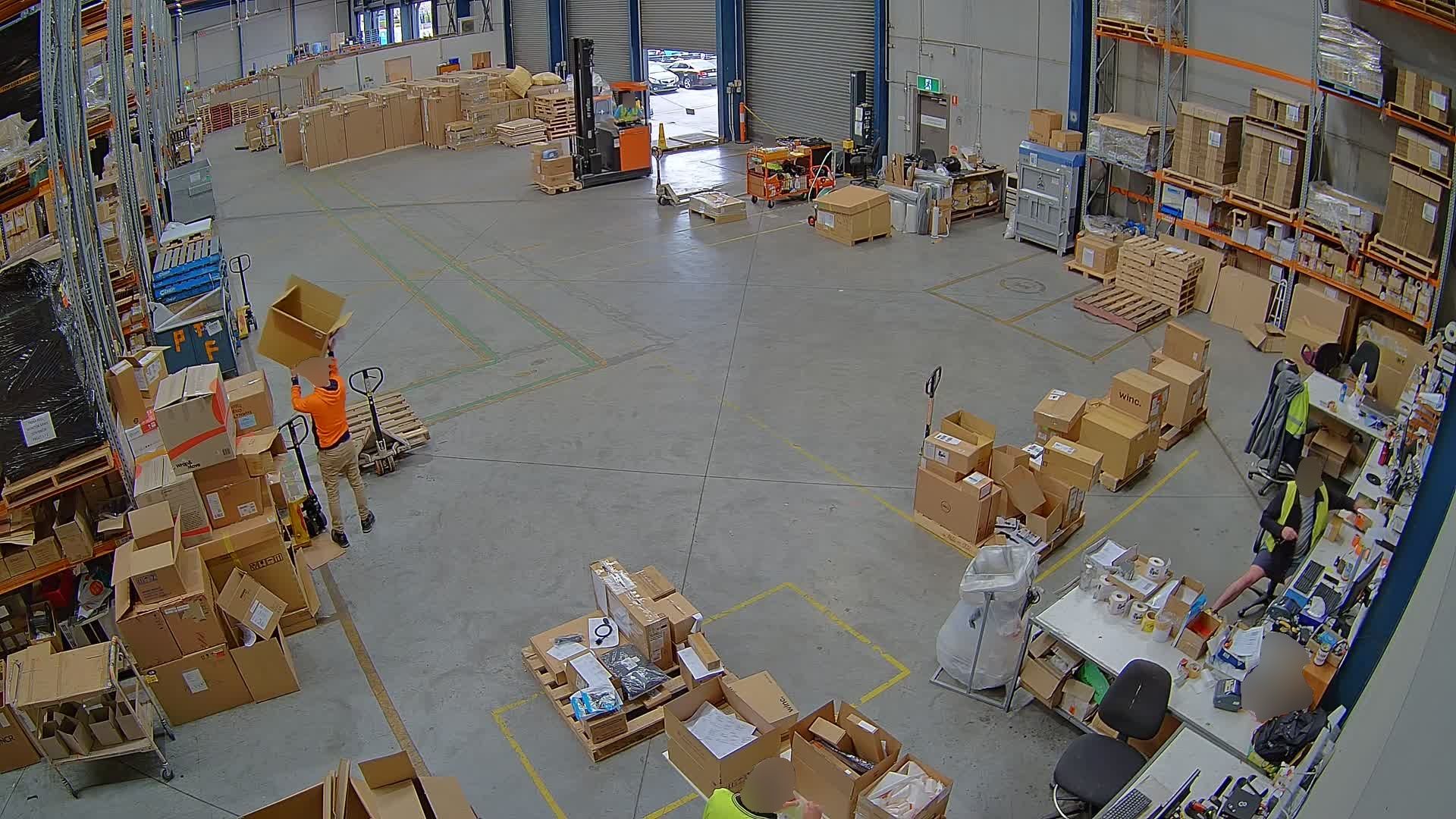}
    \caption{Input image}
    \label{fig:2.2.0005}
  \end{minipage}
\end{figure}

\subsubsection{Input}

Image (Figure \ref{fig:2.2.0005})

\subsubsection{Query}

Does this image show any workers with both hands full?

\subsubsection{Answer}

Yes, the person in the orange shirt has both hands full because they are lifting an empty box with both hands.

\subsubsection{Ground Truth}

Yes, the image shows a worker on the left with both hands full, holding a box.

\subsubsection{Score}

1.0

\newpage

\begin{figure}[h]
  \centering
  \begin{minipage}{0.49\columnwidth}
    \centering
    \includegraphics[width=0.9\columnwidth]{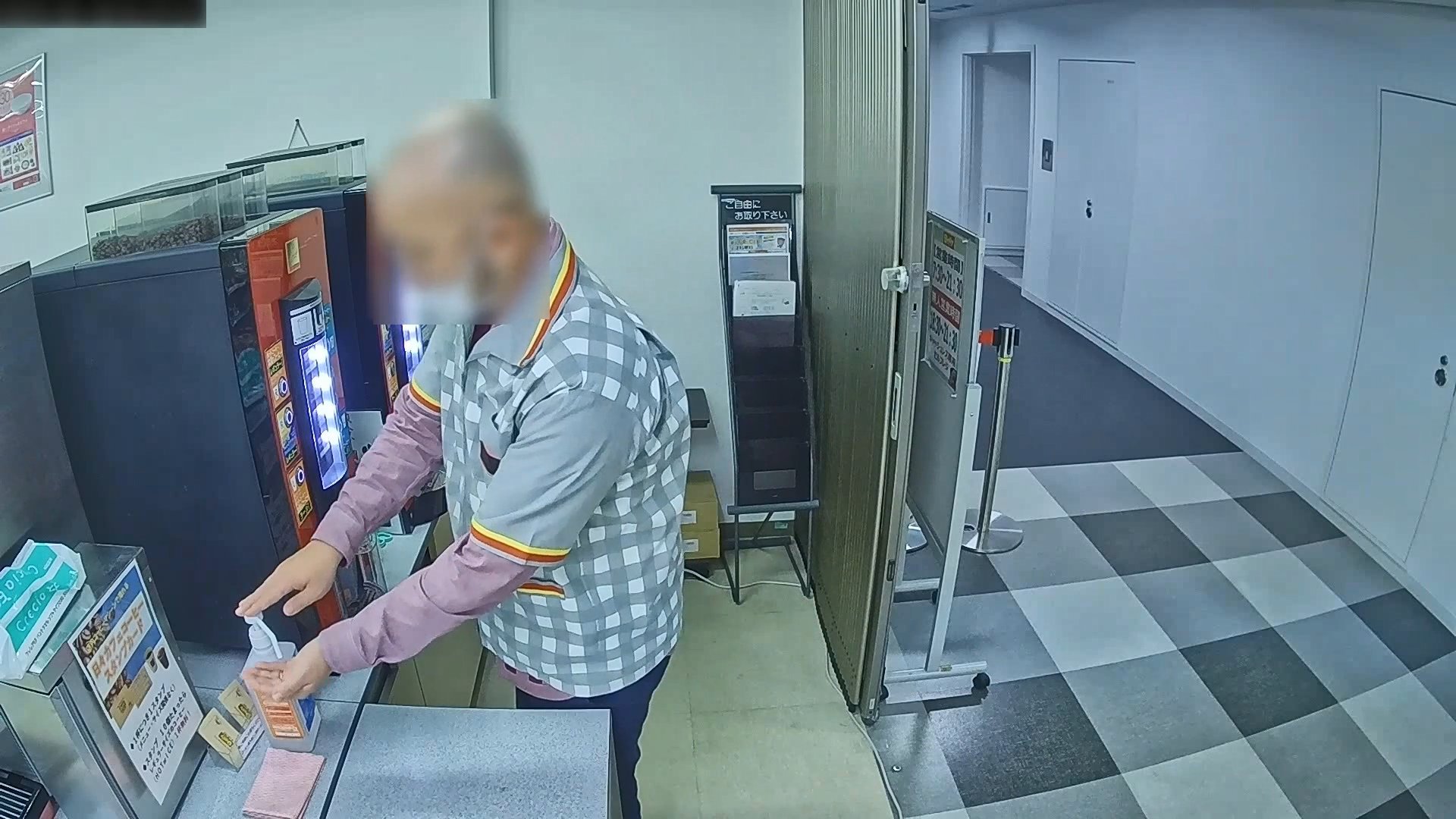}
  \end{minipage}
  \begin{minipage}{0.49\columnwidth}
    \centering
    \includegraphics[width=0.9\columnwidth]{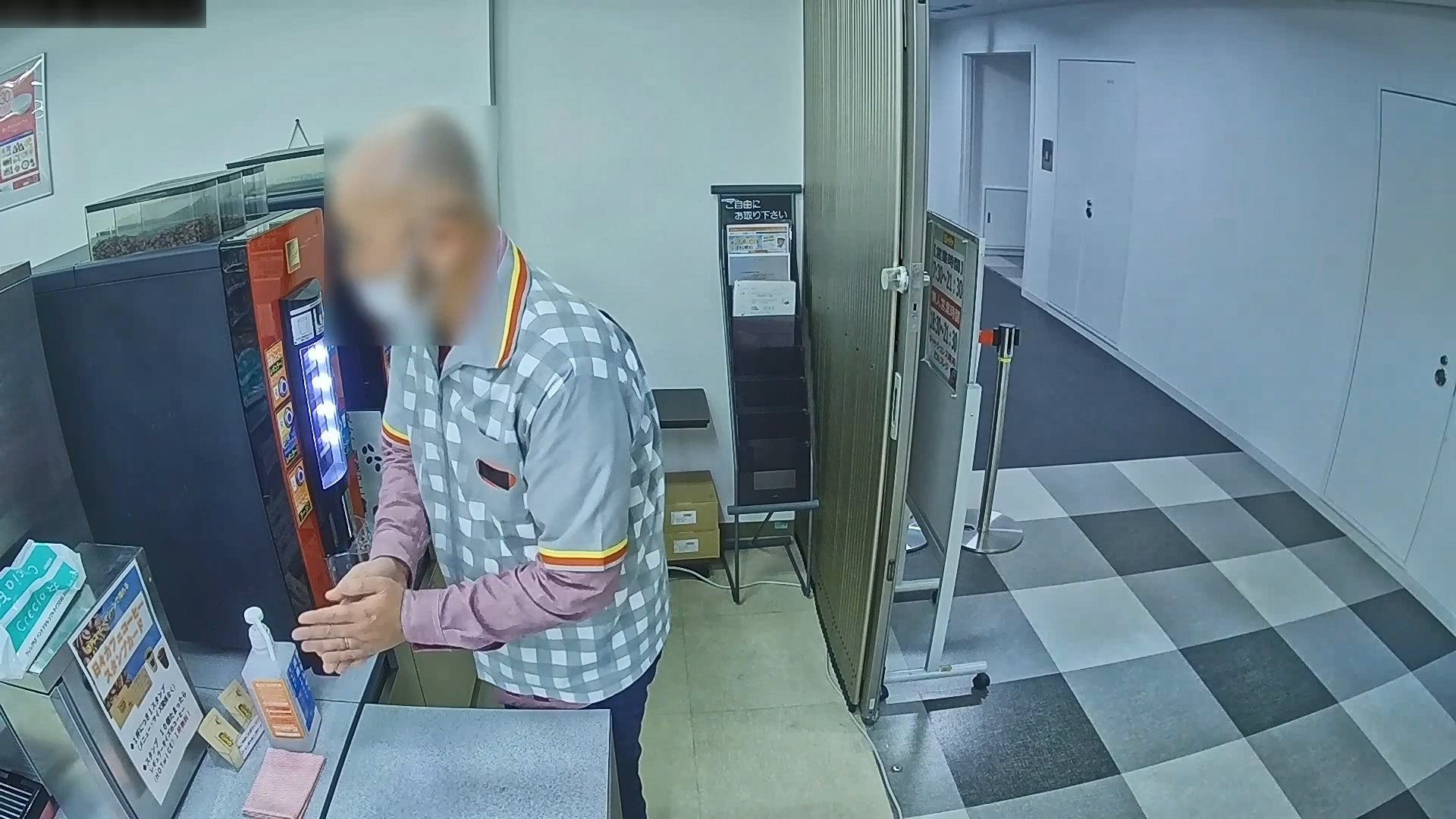}
  \end{minipage}
  \caption{Input video}
  \label{fig:1.1.2001}
\end{figure}

\subsubsection{Input}

Video (Figure \ref{fig:1.1.2001})

\subsubsection{Query}

In this video, what is the start time and what is the end time of CoffeeMaker Cleaning. The start and end times must be output from the time of the movie itself.

\subsubsection{Answer}

The specified motion start and end times are from 00:00:10 to 00:02:36.

\subsubsection{Ground Truth}

The 'CoffeeMaker Cleaning' process starts at 00:00:12 and ends at 00:02:35.

\subsubsection{Score}

0.95

\newpage

\begin{figure}[h]
  \centering
  \begin{minipage}{0.32\columnwidth}
    \centering
    \includegraphics[width=0.9\columnwidth]{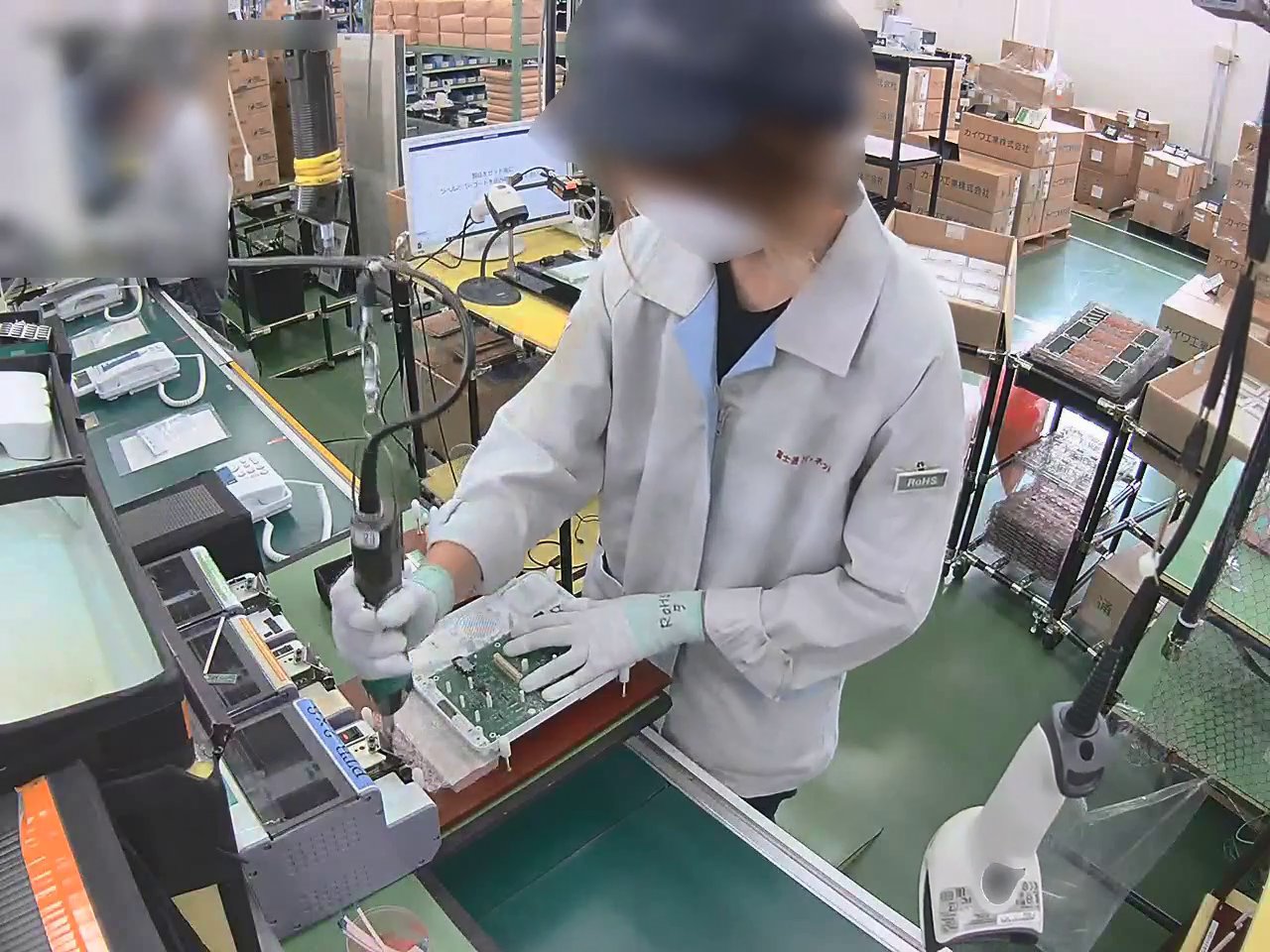}
    (a) Video 1
  \end{minipage}
  \begin{minipage}{0.32\columnwidth}
    \centering
    \includegraphics[width=0.9\columnwidth]{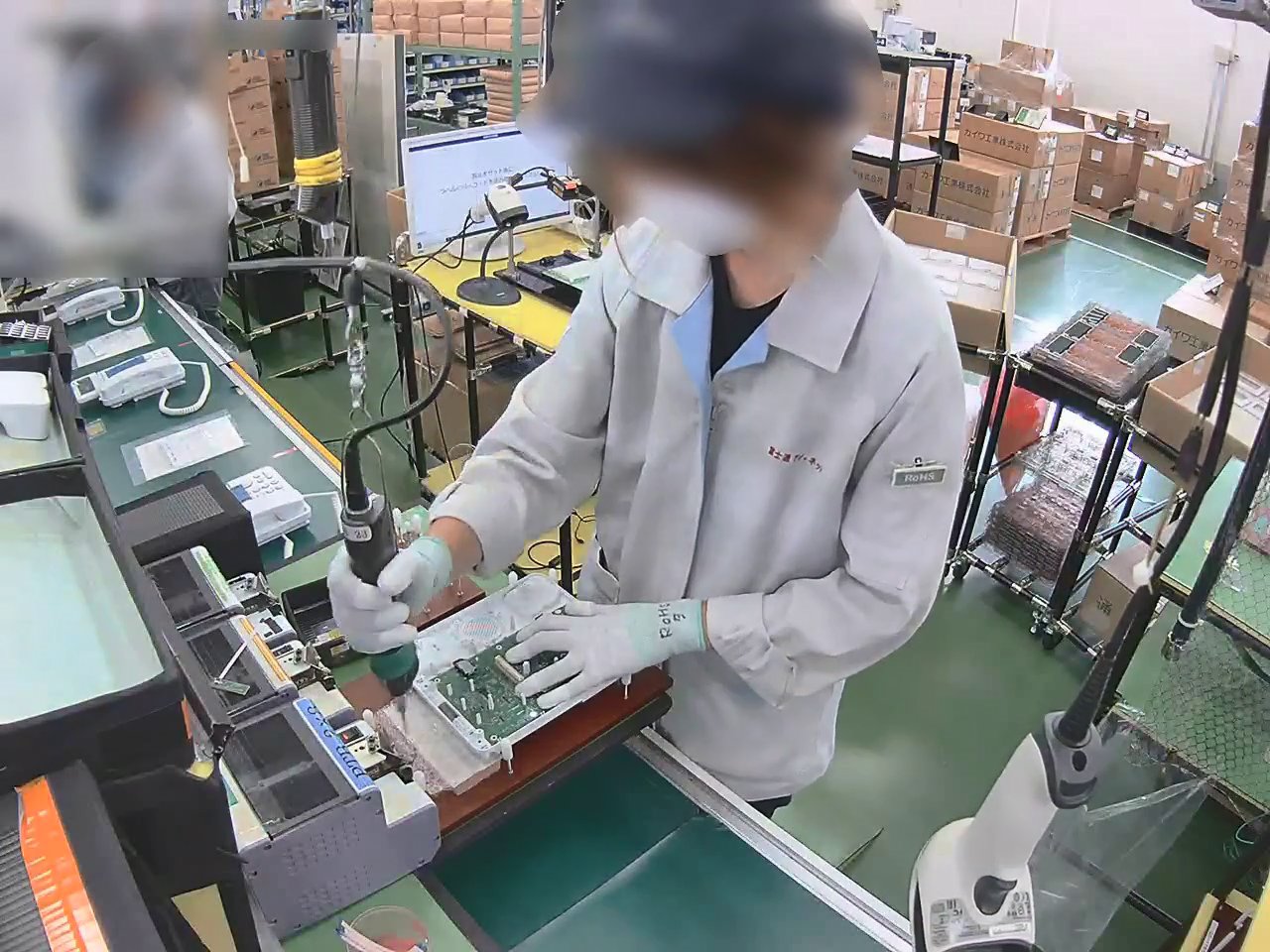}
    (b) Video 2
  \end{minipage}
  \begin{minipage}{0.32\columnwidth}
    \centering
    \includegraphics[width=0.9\columnwidth]{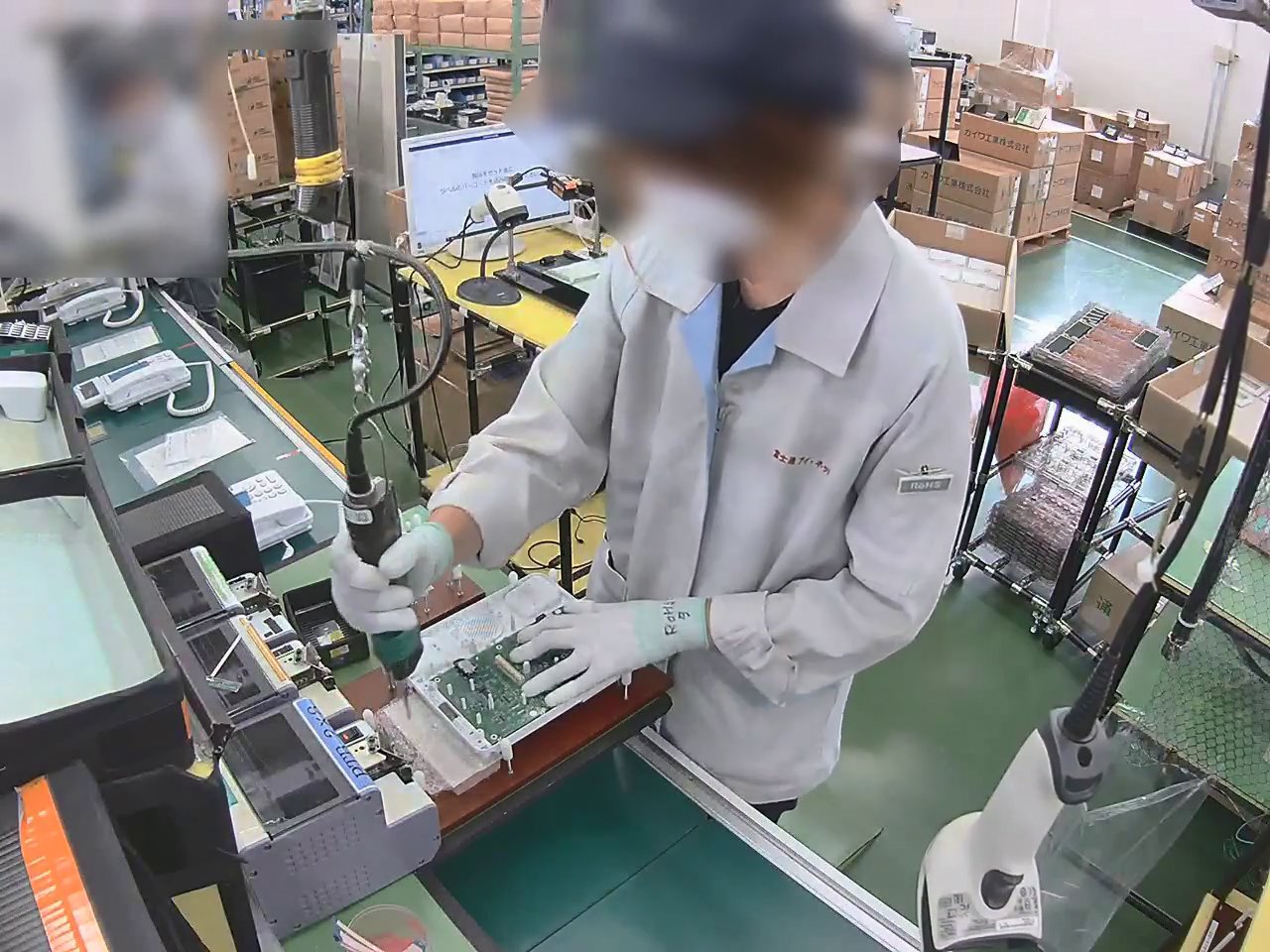}
    (c) Video 3
  \end{minipage}
  \caption{Input videos}
  \label{fig:3.4.0001}
\end{figure}

\subsubsection{Input}

Three videos (Figure \ref{fig:3.4.0001})

\subsubsection{Query}

If there were violations regarding ``the operator is using the electric screwdriver with the green label to tighten the 4 screws properly'', create a new issue for each violations with a value of ``Electric screwdriver with a green label'' for field ``Short description'', a value of filename for field ``Filename'', a value of ``violation'' for field ``Category'' and a value that must be numbered sequentially from 1110 if there are multiple incidents for field ``ID''

\subsubsection{Answer}

{"message": "All analyzed results show compliance with the instruction that the operator is using the electric screwdriver with the green label to tighten the 4 screws properly. No violations were observed, therefore no issues were created.", "status": "compliance"}

\subsubsection{Ground Truth}

{"message": "No violations regarding the use of the electric screwdriver with the green label to tighten the 4 screws properly were found in the provided videos."}

\subsubsection{Score}

1.0

\end{document}